\author{Sebastien Arnold - arnolds@usc.edu}
\title{A Greedy Algorithm to Cluster Specialists}
\date{\today}
\documentclass[12pt]{article}

\newif\iftwocolsformat
\twocolsformatfalse
\newif\iftwolinetitle
\twolinetitletrue
\newif\ifsansfont
\sansfontfalse

\usepackage[utf8]{inputenc}
\usepackage[T1]{fontenc}
\usepackage[english]{babel}
\usepackage{fancyhdr}
\usepackage{amsthm}
\usepackage{amsmath}
\usepackage{amssymb}
\usepackage{titling}
\usepackage{geometry}
\usepackage{multicol}
\usepackage{lipsum}
\usepackage{array}
\usepackage{algorithm}
\usepackage{algorithmicx}
\usepackage[noend]{algpseudocode}
\usepackage{listing}
\usepackage{graphicx}
\usepackage{longtable}
\usepackage{supertabular}
\usepackage{booktabs}
\usepackage{float}
\usepackage{hyperref}
\usepackage[export]{adjustbox}
\usepackage{breakcites}
\usepackage{color}
\usepackage{fancyvrb}
\usepackage{listings}
\usepackage{xcolor}
\usepackage{sectsty}
\usepackage{wrapfig}
\usepackage{xcolor}
\usepackage{framed}

\ifsansfont
    \allsectionsfont{\bfseries\sffamily}
    
\else
    \allsectionsfont{\bfseries\rmfamily}
\fi

\makeatletter
\let\Title\@title
\let\Author\@author
\let\Date\@date
\makeatother

\iftwocolsformat
\else
\fi
 
\makeatletter

\theoremstyle{remark}

\theoremstyle{remark}

\let\oldincludegraphics\includegraphics
\renewcommand{\includegraphics}[2][]{%
  \oldincludegraphics[#1,max width=\linewidth]{#2}}

\DefineVerbatimEnvironment{Highlighting}{Verbatim}{commandchars=\\\{\}}

\iftwolinetitle
    \def\maketitle{
        \begin{centering}
        \par\rule{\textwidth}{2pt}
        \par\hfill
        \par\textbf{\LARGE\@title}
        \par\hfill
        \par{\textit{\@author}}
        \par\hfill
        \par{\@date}
        \par\rule{\textwidth}{2pt}
        \end{centering}
    }
\else
    \def\maketitle{
        \centering
        \par\textbf{\LARGE\@title}
        \par\hfill
        \par{\@author, \@date}
        \par\hfill
        \par\hfill
        \rule{\textwidth}{3pt}
    }
\fi

\def\maketitle{
    \begin{centering}
    \par\rule{\textwidth}{2pt}
    \par\hfill
    \par\textbf{\LARGE\@title}
    \par\hfill
    \par{\textit{\@author}}
    \par\hfill
    \par{\@date}
    \par\rule{\textwidth}{2pt}
    \end{centering}
}

\begin{document}
\thispagestyle{empty}
\maketitle
\hfill

\abstract{
Several recent deep neural networks experiments leverage the
generalist-specialist paradigm for classification. However, no formal study
compared the performance of different clustering algorithms for class
assignment. In this paper we perform such a study, suggest slight modifications
to the clustering procedures, and propose a novel algorithm designed to
optimize the performance of of the specialist-generalist classification system.
Our experiments on the CIFAR-10 and CIFAR-100 datasets allow us to investigate
situations for varying number of classes on similar data. We find that our
\emph{greedy pairs} clustering algorithm consistently outperforms other
alternatives, while the choice of the confusion matrix has little impact on the
final performance.
}

\section{Introduction}\label{introduction}

Designing an efficient classification system using deep neural networks
is a complicated task, which often use a multitude of models arranged in
ensembles. \cite{galaxy}, \cite{vgg} These ensembles often lead to
state-of-the-art results on a wide range of different tasks such as
image classification \cite{inception}, speech recognition
\cite{deepspeech2}, and machine translation \cite{seq2seq}. The
models are trained independently and in parallel, and different
techniques can be used to merge their predictions.

\begin{figure}[b]
\centering
\includegraphics[width=\textwidth]{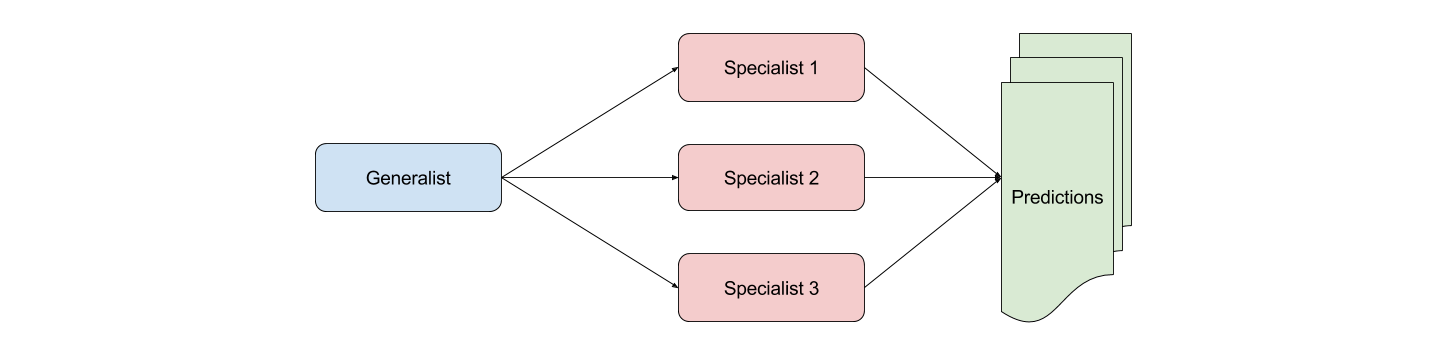}
\caption{An example of specialist architecture with three specialists.}
\label{fig:1}
\end{figure}

A more structured alternative to ensembling is the use of the
specialist-generalist framework. As described by \cite{bochereau1990}, a
natural analogy can be drawn from the medical field; a patient first
consults a general practitioner who provides an initial diagnosis which
is then refined by one or several specialists. In the case of
classification, the doctors are replaced by neural networks and the
final prediction is a combination of the specialists' outputs, and may
or may not include the generalist's take.

In recent years, generalist and specialists have been studied under
different circumstances. \cite{darkknowledge} used specialists to create
an efficient image classifier for a large private dataset. The final
predictions of the specialists were then used to train a reduced
classifier that achieved performance similar to the whole ensemble.
\cite{emonets} describe a multimodal approach for emotion recognition in
videos, based on specialists. Maybe closer to our work,
\cite{wardefarley} added ``auxiliary heads'' (acting as specialists) to
their baseline network, using the precomputed features for both
classification and clustering. They also underlined one of the main
advantages of using specialists; a relatively low (and parallelizable)
additional computational cost for increased performance.

\section{Clustering Algorithms}\label{clustering-algorithms}

In the generalist-specialist framework, each class is assigned to one or more
specialists. This assignment is usually done by clustering the classes into
non-overlapping sets. The goal of the clustering procedure is to optimize the
generalist vs specialist accuracy trade-off. In that sense, the algorithm must
carefully balance the classification performance of the generalist over the set
of clusters and the specialist performance within those clusters. In the case
of overlapping clusters, an additional weighting parameter for specialists can be added at
inference time.

In order to assign classes to the specialist networks, we compare
several clustering algorithms on the confusion matrix of the outputs of
the generalist. This confusion matrix is computed on a held-out
partition of the dataset. Following previous works, we started by
considering two baseline clustering algorithms, namely Lloyd's K-Means
algorithm and Spectral clustering, according to the formulation of
\cite{spectral}. In addition to those baseline algorithms, we evaluate
the performance of two novel procedures specifically designed to improve
the generalist-specialist paradigm. Those algorithms are described in
the following paragraphs, and pseudo code is given in the Appendix.

We also experimented with different ways of building the confusion matrix.
Besides the usual way of constructing a confusion matrix by accumulating all
predictions for each classes (denoted here as \emph{standard}), we tried three
alternatives:

\begin{itemize}
\itemsep1pt\parskip0pt\parsep0pt
\item
  \emph{softsum}: for each prediction, we use the raw model output
  instead of the one-hot multi-class output,
\item
  \emph{softsum pred}: just like \emph{softsum}, but only add the
  prediction output to the confusion matrix, if the class was correctly
  predicted,
\item
  \emph{softsum not pred}: like to \emph{softsum pred}, but only if
  the prediction output was incorrectly predicted.
\end{itemize}

As discussed in later sections, the influence of the confusion matrix is
minimal. Nonetheless we include them for completeness purposes.

Both of our clustering algorithms further modify the confusion matrix
$A$ by computing $CM = \textbf{A}^\top + \textbf{A}$, which symmetrizes
the matrix. We define the entries of the matrix to be the
\emph{animosity score} between two classes; given classes \emph{a} and
\emph{b}, their animosity score is found at $CM_{a, b}$. We then
initialize each cluster with non-overlapping pairs of classes yielding
maximal animosity score. Finally, we greedily select the next classes to
be added to the clusters, according to the following rules:

\begin{itemize}
\item
  In the case of \emph{greedy single} clustering, a single class
  maximizing the overall animosity score is added to the cluster
  yielding the largest averaged sum of animosity towards this class.
  This partitions the classes in clusters, building on the intuition
  that classes that are hard to distinguish should be put together.
\item
  In the case of \emph{greedy pairs} clustering, we follow the same
  strategy as in \emph{greedy single} clustering but act on pair of
  classes instead of single classes. In this case we allow the clusters
  to overlap, and one prediction might include the opinion of several
  specialists.
\end{itemize}

This process is repeated until all classes have been assigned to at
least one cluster.

\section{Experiments}\label{experiments}

We investigate the performance of the aforementioned algorithms on the
CIFAR-10 and CIFAR-100 datasets (\cite{cifar}). Both datasets contain
similar images, partitioned in 45'000 train, 5'000 validation, and
10'000 test images. They contain 10 and 100 classes respectively. For
both experiments we train the generalist network on the train set only,
and use the validation set for clustering purposes. As we are interested
in the clustering performance we do not augment nor pre-process the
images. Note that when trained on the horizontally flipped training and
validation set our baseline algorithm reaches 10.18\% and 32.22\%
misclassification error respectively, which is competitive with the
current state-of-the-art presented in \cite{allcnn}.

Following \cite{binaryconnect}, the baseline network is based on the
conclusions of \cite{vgg} and uses three pairs of batch-normalized
convolutional layers, each followed by a max-pooling layer, and two
fully-connected layers. The same model is used for specialists, whose
weights are initialized with the trained weights of the generalist.
\footnote{The code for these experiments is freely available online at
  \href{http://www.github.com/seba-1511/specialists}.}
One major departure from the work of \cite{darkknowledge} is that our
specialists are predicting over the same classes as the generalist,
i.e.~we do not merge all classes outside of the cluster into a unique
one. With regards to the generalist, a specialist is only biased towards
a subset of the classes, since it has been fine-tuned to perform well on
those ones.

\subsection{CIFAR-10}\label{cifar-10}

\begin{longtable}[c]{@{}lllll@{}}
\toprule\addlinespace
Results & standard & soft sum & soft sum pred & soft sum not pred
\\\addlinespace
\midrule\endhead
spectral & (0.7046, 2) & (0.7719, 2) & (0.6989, 2) & (0.706, 2)
\\\addlinespace
greedy singles & (0.5873, 2) & (0.5049, 2) & (0.5139, 3) & (0.5873, 2)
\\\addlinespace
kmeans & (0.8202, 2) & (0.8202, 2) & (0.8202, 2) & (0.8202, 2)
\\\addlinespace
greedy pairs & (0.8835, 2) & (0.8835, 2) & (0.8727, 3) & (0.8835, 2)
\\\addlinespace
\bottomrule
\addlinespace
\caption{Experiment results for CIFAR-10}
\end{longtable}

For CIFAR-10 experiments, we considered up to five clusters, and all of the
possible combinations of confusion matrix and clustering algorithms. The
results for this experiments are reported in Table 1. For each clustering
algorithm and confusion matrix type we report first the obtained accuracy, and
then the number of clusters to reach it.


Interestingly, the choice of confusion matrix has only a limited impact
on the overall performance, indicating that the emphasis should be put
on the clustering algorithm. We notice that clustering with greedy pairs
consistently yields better scores. However none of the specialist
experiments is able to improve on the baseline, suggesting that
specialists might not be the framework of choice when dealing with a
small number of classes.

\subsection{CIFAR-100}\label{cifar-100}

For CIFAR-100 we performed the exact same experiment as for CIFAR-10 but used
more specialists, the largest experiments involving 28 clusters. The results
are shown in Table 2. Again, we report the obtained accuracy and the number of
clusters for each clustering algorithm and confusion matrix type.


\begin{longtable}[c]{@{}lllll@{}}
\toprule\addlinespace
Results & standard & soft sum & soft sum pred & soft sum not pred
\\\addlinespace
\midrule\endhead
spectral & (0.5828, 2) & (0.5713, 2) & (0.5755, 2) & (0.5795, 3)
\\\addlinespace
greedy singles & (0.3834, 2) & (0.3733, 2) & (0.3803, 2) & (0.3551, 2)
\\\addlinespace
kmeans & (0.5908, 2) & (0.5618, 2) & (0.5820, 3) & (0.5876, 2)
\\\addlinespace
greedy pairs & (0.6141, 6) & (0.5993, 6) & (0.6111, 6) & (0.607, 6)
\\\addlinespace
\bottomrule
\addlinespace
\caption{Experiment results for CIFAR-100}
\end{longtable}

Similarly to CIFAR-10, we observe that greedy pairs clustering
outperforms the other clustering techniques, and that the different
types of confusion matrix have a limited influence on the final score.
We also notice that fewer clusters tend to work better. Finally, and
unlike the results for CIFAR-10, some of the specialists are able to
improve upon the generalist, which confirms our intuition that
specialists are better suited to problems involving numerous output
classes.

We suggest the following explanation for the improved performance of
greedy pairs is the following. Allowing clusters to overlap leads to the
assignment of difficult classes to multiple specialists. At inference
time, more networks will influence the final prediction which is
analogous to building a larger ensemble for difficult classes.

\section{Conclusion and Future Work}\label{conclusion-and-future-work}

We introduce a novel clustering algorithm for the specialist-generalist
framework, which is able to consistently outperform other techniques. We
also provide a preliminary study of the different factors coming into
play when dealing with specialists, and conclude that the choice of
confusion matrix from our proposed set only has little impact on the
final classification outcome.

Despite our encouraging results with clustering techniques, no one of
our specialists-based experiments came close to compete with the
generalist model trained on the entire train and validation set. This
was a surprising outcome and we suppose that this effect comes from the
size of the datasets. In both cases, 5'000 images corresponds to 10\% of
the original training set and removing that many training examples has a
drastic effect on both generalists and specialists. All the more so
since we are not using any kind of data augmentation techniques, which
could have moderated this downside. An obvious future step is to
validate the presented ideas on a much larger dataset such as
Imagenet \cite{imagenet} where splitting the train set would not hurt the train
score as much.

\subsubsection{Acknowledgments}

We would like to thank Greg Ver Steeg, Gabriel Pereyra, and Pranav Rajpurkar for their comments and advices. We also thank Nervana Systems
for providing GPUs as well as their help with their deep learning
framework.

\section{Appendix}\label{appendix}


\begin{algorithm}[H]
    \caption{Greedy Pairs Clustering}
    \label{greedy_pairs}
    \begin{algorithmic}[1] 
        \Procedure{GreedyPairs}{$M,N$} \Comment{Confusion matrix M, number of clusters N}
            \State $M\gets M + M^T$
            \State Initialize N clusters with non-overlapping pairs maximizing the entries of M.
            \While{every class has not been assigned}
                \State Get the next pair $(a, b)$ maximizing the entry in M
                \State cluster = $\underset{\text{c in clusters}}{\mathrm{argmin}}$(Animosity(a, c) + Animosity(b, c))
                \State Assign(cluster, a, b)
            \EndWhile\label{euclidendwhile}
            \State \textbf{return} clusters
        \EndProcedure
    \end{algorithmic}
\end{algorithm}





\begin{thebibliography}{99.}%


    \bibitem{bochereau1990}Bochereau, Laurent, and Bourgine, Paul. A Generalist-Specialist Paradigm for
Multilayer Neural Networks. Neural Networks, 1990.


\bibitem{binaryconnect}Courbariaux, Matthieu, Bengio, Yoshua, and David, Jean-Pierre. BinaryConnect:
Training Deep Neural Networks with Binary Weights during Propagations. NIPS,
2015.


\bibitem{galaxy}Dieleman, Sander, Willett, Kyle W., and Dambre, Joni. Rotation-invarient
convolutional neural networks for galaxy morphology prediction. Oxford Journals,
2015.


\bibitem{deepspeech2}Hannun, Awni, Case, Carl, Casper, Jared, Catanzaro, Bryan, Diamos, Greg, Elsen,
Erich, Prenger, Ryan, Satheesh, Sanjeev, Sengupta, Shubho, Coates, Adam, and Ng,
Andrew Y. Deep Speach: Scaling up end-to-end speech recognition. Arxiv Preprint,
2014.


\bibitem{darkknowledge}Hinton, Geoffrey E., Vinyals, Oriol, and Dean, Jeff. Distilling th Knowledge in
a Neural Network. NIPS 2014 Deep Learning Workshop.


\bibitem{emonets}Kahou, Samira Ebrahimi, Bouthiller, Xavier, Lamblin, Pascal, Gulcehre, Caglar,
Michalski, Vincent, Konda, Kishore, Jean, Sébastien, Froumenty, Pierre, Dauphin,
Yann, Boulanger-Lewandowski, Nicolas, Ferrari, Raul Chandias, Mirza, Mehdi,
Warde-Farley, David, Courville, Aaron, Vincent, Pascal, Memisevic, Roland, Pal,
Christopher, and Bengio, Yoshua. EmoNets: Multimodal deep learning approaches
for emation recofnition in video. Journal on Mutlimodal User Interfaces, 2015.


\bibitem{cifar}Krizhevsky, Alex. Learning Multiple Layers of Features from Tiny Images. 2009.


\bibitem{spectral}Ng, Andrew Y., Jordan, Micheal I., Weiss, Yair. On spectral clustering: Analysis
and an algorithm. NIPS 2002.


\bibitem{imagenet}Russakovsky, Olga, Deng, Jia, Su, Hao, Krause, Jonathan, Satheesh, Sanjeev, Ma,
Sean, huang, Zhiheng, Karpathy, Andrej, Khosla, Aditya, Bernstain, Michael,
Berg, Alexander C., and Fei-Fei, Li. ImageNet Large Scale Visual Recognition
Challenge. International Journal of Computer Vision, 2015.


\bibitem{vgg}Simonyan, Karen and Zisserman, Andrew. Very Deep Convolutional Networks for
Large-Scale Image Recognition. International Conference on Learning
Representations, 2015.


\bibitem{allcnn}Springenberg, Jost Tobias, Dosovitskiy, Alexey, Brox, Thomas, and Riedmiller,
Martin. Striving for Simplicity: The All Convolutional Net. International
Conference on Learning Representations Workshop, 2015.


\bibitem{seq2seq}Sutskever, Ilya, Vinyals, Oriol, and Le, Quoc V. Sequence to Sequence Learning with
Neural Networks. Arxiv Preprint, 2014.


\bibitem{inception}Szegedy, Christian, Liu, Wei, Jia, Yangqing, Sermanet, Pierre, Reed, Scott,
Anguelov, Dragomir, Erhan, Dumitru, Vanhoucke, Vincent, and Rabinovich, Andrew.
Going deeper with convolutions. Arxiv Preprint, 2014.


\bibitem{wardefarley}Warde-Farley, David, Rabinovich, Andrew, and  Anguelov, Dragomir. Self-Informed
Neural Networks Structure Learning. International Conference on Representations
Learning, 2015.
\end{thebibliography}

\end{document}